\documentclass[pdflatex,sn-basic,Numbered,iicol]{sn-jnl}
\usepackage{booktabs,tabularx,array,amsmath,enumitem,xcolor,graphicx,tikz}
\usetikzlibrary{arrows.meta,positioning,fit,backgrounds}
\usepackage{xurl,capt-of,etoolbox,needspace}
\makeatletter
\patchcmd{\@maketitle}{\vskip21pt}{\vskip0pt}{}{}
\patchcmd{\@maketitle}{\vskip20pt}{\vskip10pt}{}{}
\patchcmd{\@maketitle}{\vskip20pt}{\vskip10pt}{}{}
\patchcmd{\@maketitle}{\vskip24pt}{\vskip12pt}{}{}
\patchcmd{\@maketitle}{\vskip24pt}{\vskip12pt}{}{}
\patchcmd{\@maketitle}{\vskip36pt}{\vskip20pt}{}{}
\patchcmd{\@maketitle}{Corresponding author(s). E-mail(s):}{Corresponding author(s):}{}{}
\makeatother
\hypersetup{hidelinks}
\definecolor{ink}{HTML}{16354A}
\definecolor{accent}{HTML}{287D8E}
\definecolor{pale}{HTML}{EEF5F7}
\setlist{nosep,leftmargin=*}

\title[Putting a Robot on a Phone Call]{Putting a Robot on a Phone Call: Layered Signaling, Bounded Occupancy and Turn Arbitration over WebRTC}
\author*[1]{\fnm{Zheyuan} \sur{Zhang}}\email{1120230155@bit.edu.cn}
\author*[2]{\fnm{Zaide} \sur{Zhang}}\email{zhangzaide7@gmail.com}
\affil[1]{\orgname{Beijing Institute of Technology}, \orgaddress{\city{Beijing}, \country{China}}}
\affil[2]{\orgname{NovaXBot}, \orgaddress{\city{Beijing}, \country{China}}}
\makeatletter
\patchcmd{\auaddress}{.\par}{\par}{}{\PackageError{author-format}{First affiliation punctuation patch failed}{}}
\patchcmd{\auaddress}{.\par}{\par}{}{\PackageError{author-format}{Second affiliation punctuation patch failed}{}}
\def\corrauthemail{\textcolor{blue}{1120230155@bit.edu.cn};\ \textcolor{blue}{zhangzaide7@gmail.com}}
\makeatother
\abstract{We put a Thor robot on a browser voice call, with speech sent from Safari, answers played through the robot's loudspeaker and transcripts returned to the phone. We separate HTTPS signaling, relayed WebRTC media and speech generation into independently observable routes. The browser prefetches relay credentials and allows a 500 ms window after its first relay candidate. The backend starts speech and WebRTC-answer preparation concurrently within a 55 s deadline. A session slot and a non-blocking speaker lock coordinate calls with local voice access. A 2 s gate gives the controller an opportunity to suppress chat before playback. Fifteen of sixteen first dials reached ready-to-speak, comprising seven of eight Wi-Fi calls with a median of 5.57 s and eight of eight cellular calls with a median of 6.76 s. All twenty scored occupancy and recovery scenarios admitted a later call. During five 60 s outages, the backend released the line while the phone remained offline. Across 40 controlled replays, enabling the gate suppressed chat when the controller decided at 1 s. At 3 s, chat and controller feedback both played in sequence, with no overlapping output in any trace. Two additional boundary traces exposed a receipt naming a later feedback item and a new call's query returning its predecessor's feedback. These measurements cover application events on one platform and simulated controller decisions, with acoustic output and task correctness outside their scope.
}
\begin{document}
\maketitle
\section{Introduction}
A person taps Call on an iPhone. The answer comes from the speaker beside Thor, while the caller reads the same words on the phone. That arrangement puts the caller's microphone, a public web entry, a speech service and a shared loudspeaker on one conversational path. It also gives a task controller and a conversational model an opportunity to answer the same utterance.

We need to decide when the line is ready, who holds the speaker and which answer gets the next turn. A loaded page establishes access to the web entry. A working call additionally needs relayed media and a live speech connection. Once those connections exist, local voice and another remote caller still compete for the physical output device. Even with one caller admitted, a delayed controller decision may arrive after chat has begun speaking.

We build a browser-to-robot voice path around these boundaries. Separate signaling, media and speech routes expose connection state at each layer. Credential prefetch and concurrent backend setup overlap independent waits. A session slot and an operating-system speaker lock coordinate remote and local access, while a bounded gate lets the controller claim a turn before chat reaches playback. The same path reports text and software playback events to the caller.

We evaluated first-dial readiness over Wi-Fi and cellular access, recovery under four disturbances, and output ordering under controlled controller timings. Two boundary traces locate feedback-association errors within a call and across calls.

\section{Related Work}
\subsection{Making a robot remotely accessible}
Mobile telepresence connects a remote operator with people at the robot's physical site \cite{kristoffersson,desai2011,tsui2011}. Research on socially interactive robots and HRI situates that access within ongoing human activity \cite{fong2003,goodrich}. Cloud robotics and web interfaces extend the computation and interfaces available remotely \cite{kehoe2015,toris}. We address access at the loudspeaker, where remote calls and a local voice process contend for the same device.

\subsection{Allocating a speaking turn}
Conversation analysis describes how participants allocate the next turn \cite{sacks1974}. Spoken-dialogue research examines timing and multimodal cues around a turn boundary \cite{skantze2014,skantze2021}. Engagement research follows people as they enter or leave an interaction \cite{bohus2009,bohus2014}. We apply an explicit deadline to the software sources competing after the caller yields the floor.

\subsection{Identifying the source of a robot voice}
Physical embodiment affects how people encounter a robot \cite{kidd2004,li2015}. Studies of voice interfaces describe how users interpret replies and revise expectations after misunderstandings \cite{luger2016,clark2019,porcheron2018}. Language-grounded robotics and language-model dialogue provide several routes from an utterance to an answer \cite{tellex2020,furchat2023}. We preserve the submitting component's identity through the response path so that a controller status and a generated answer remain distinguishable in the record.

\subsection{Choosing the evaluation target}
HRI measures depend on the question being asked \cite{steinfeld}. Usability ratings and perceived-intelligence measures address a user's experience \cite{bangor2008,bartneck2009}. We evaluate application events at connection, resource and output boundaries, using the audio fixture's reference mapping to check the identity reported by the player.

\section{Design Problem and Three Principles}
\subsection{Four commitments in a call}
We fix four things for each call. The session identifies the connection whose lifecycle is being managed. Speaker ownership identifies the process allowed to write to the physical device. Turn ownership identifies the subsystem permitted to answer a particular utterance. Evidence identifies what an exported event establishes about the call. A controller may take a turn while the same caller continues to hold the speaker.

These commitments have different lifetimes. A task callback may outlive its call, and a feedback submission may precede completion of earlier audio. Identifiers must survive asynchronous delivery and observations must retain their source.

\subsection{Three design principles}
Independently failing connection paths must remain separately observable. Contention for the line, speaker and next turn requires an explicit owner and a bounded wait. Asynchronous output must retain its originating call and turn, while each observation identifies the claim it supports.

\subsection{Seven evidence levels}
A controller may report task completion while its spoken feedback waits in the queue. Table~\ref{tab:evidence} assigns those observations to their corresponding claims. We use the levels to select a measurement for each design choice and to keep an answer-area message's source visible during analysis.

\begin{table*}[t]
\centering\small
\caption{Applying seven evidence levels to a robot call. Each row identifies what one observation establishes.}\label{tab:evidence}
\begin{tabularx}{\linewidth}{@{}p{.28\linewidth}X@{}}\toprule
Observation & Claim supported by the record \\\midrule
Input transcript & The recognizer returned these words for the input. \\
Generated response text & The service produced this answer text. \\
Audio queued & The playback process accepted audio for scheduling. \\
Software playback receipt & The player reported completion for the identified audio. \\
Controller feedback & The controller reported the indicated task state. \\
Acoustic recording & A microphone captured sound at its location. \\
Listener or task observation & A listener interpreted the utterance, or an independent check verified the physical outcome, under the stated test conditions. \\\bottomrule
\end{tabularx}
\end{table*}

\section{System}
\subsection{Separating signaling, media and speech}
We need to inspect the connection beyond the page-load event. A browser may reach the public entry while media negotiation or speech generation encounters a separate failure. The web entry also has its own lifecycle, so restarting it requires us to inspect the established call's downstream state.

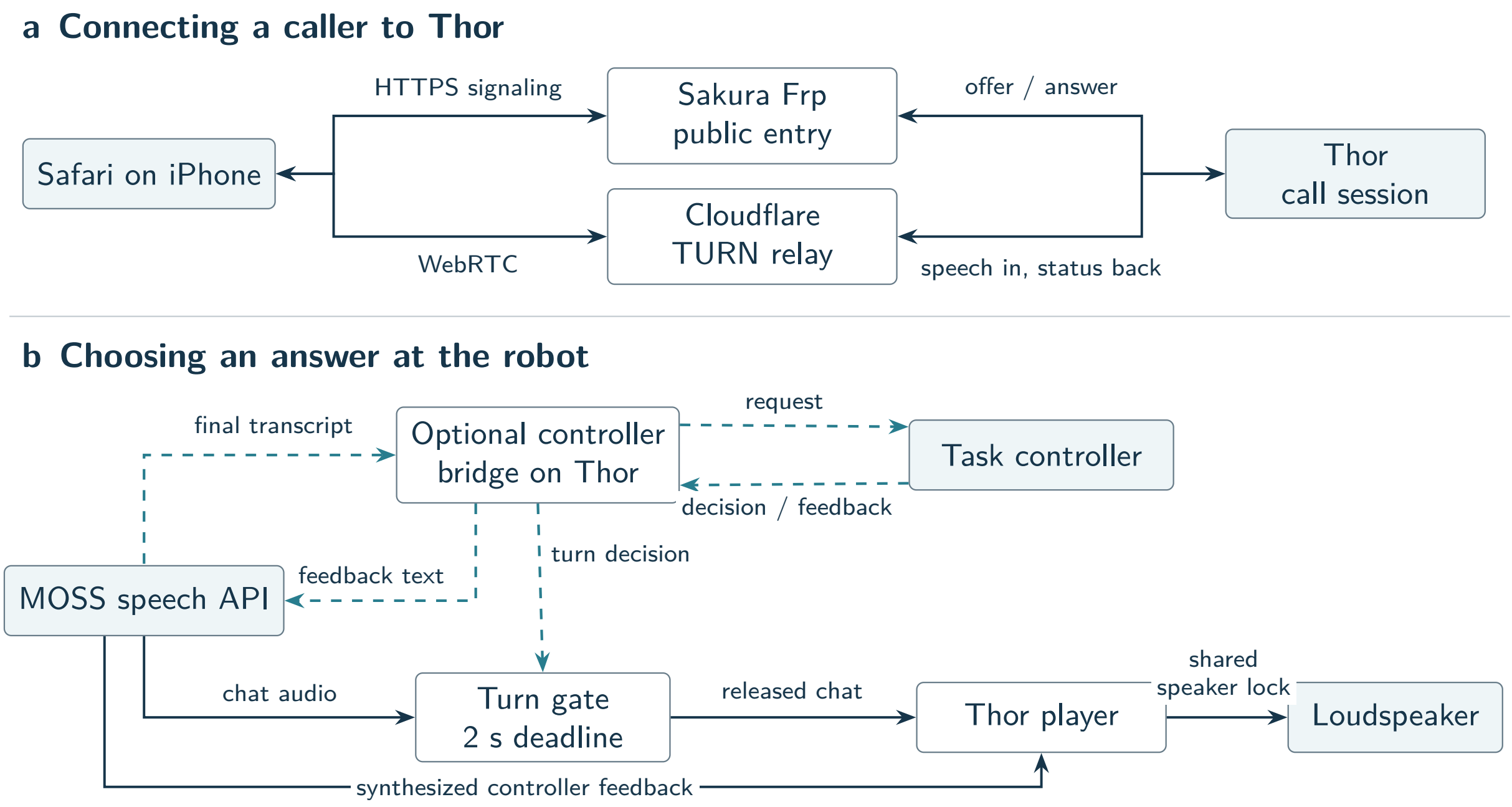
\begin{figure*}[!t]
\centering
\resizebox{\linewidth}{!}{%
\begin{tikzpicture}[
 x=1cm,y=1cm,font=\sffamily\small,
 box/.style={draw=ink!65,rounded corners=2pt,fill=white,align=center,
             inner sep=4pt,minimum height=.70cm,text=ink},
 endpoint/.style={box,fill=pale},
 link/.style={-{Stealth[length=2mm]},line width=.7pt,draw=ink},
 both/.style={{Stealth[length=2mm]}-{Stealth[length=2mm]},line width=.7pt,draw=ink},
 ctrl/.style={-{Stealth[length=2mm]},line width=.75pt,draw=accent,dashed},
 flowlabel/.style={align=center,font=\sffamily\footnotesize,text=ink,fill=white,inner sep=1.6pt},
 heading/.style={anchor=west,font=\sffamily\bfseries,text=ink}
]
\node[heading] at (-.35,6.9) {a\enspace Connecting a caller to Thor};
\node[endpoint,minimum width=2.5cm] (phone) at (1.05,5.45) {Safari on iPhone};
\node[box,minimum width=2.9cm] (web) at (7.1,6.03) {Sakura Frp\\public entry};
\node[box,minimum width=2.9cm] (relay) at (7.1,4.82) {Cloudflare\\TURN relay};
\node[endpoint,minimum width=2.6cm] (call) at (13.15,5.45) {Thor\\call session};
\draw[both] (phone.east) -- ++(.58,0) |- (web.west);
\draw[both] (phone.east) -- ++(.58,0) |- (relay.west);
\draw[both] (web.east) -- ++(2.45,0) |- (call.west);
\draw[both] (relay.east) -- ++(2.45,0) |- (call.west);
\node[flowlabel] at (4.25,6.3) {HTTPS signaling};
\node[flowlabel] at (4.25,4.55) {WebRTC};
\node[flowlabel] at (10.0,6.31) {offer / answer};
\node[flowlabel] at (10.0,4.49) {speech in, status back};
\draw[ink!25] (-.35,4.02) -- (14.7,4.02);
\node[heading] at (-.35,3.60) {b\enspace Choosing an answer at the robot};
\node[endpoint,minimum width=2.6cm] (moss) at (1.0,1.17) {MOSS speech API};
\node[box,minimum width=2.65cm] (bridge) at (4.95,2.63) {Optional controller\\bridge on Thor};
\node[endpoint,minimum width=2.65cm] (controller) at (10.0,2.63) {Task controller};
\node[box,minimum width=2.55cm] (gate) at (5.0,0) {Turn gate\\2 s deadline};
\node[box,minimum width=2.5cm] (player) at (10.0,0) {Thor player};
\node[endpoint,minimum width=2.15cm] (speaker) at (13.55,0) {Loudspeaker};
\draw[ctrl] (1.0,1.54) |- (bridge.west);
\node[flowlabel] at (2.3,2.91) {final transcript};
\draw[ctrl] (bridge.12) -- (controller.168);
\draw[ctrl] (controller.192) -- (bridge.-12);
\node[flowlabel] at (7.42,3.14) {request};
\node[flowlabel] at (7.45,2.09) {decision / feedback};
\draw[ctrl] (bridge.south) -- (gate.north);
\node[flowlabel,anchor=west] at (5.02,1.65) {turn decision};
\draw[ctrl] (bridge.218) |- (moss.east);
\node[flowlabel] at (3.28,1.43) {feedback text};
\draw[link] (moss.south) |- (gate.west);
\node[flowlabel] at (2.36,.25) {chat audio};
\draw[link] (gate.east) -- (player.west);
\node[flowlabel] at (7.5,.27) {released chat};
\draw[link] (moss.222) -- ++(0,-1.51) -| (player.south);
\node[flowlabel] at (4.82,-.72) {synthesized controller feedback};
\draw[link] (player.east) -- (speaker.west);
\node[flowlabel] at (11.83,.43) {shared\\speaker lock};
\end{tikzpicture}}
\caption{Signaling, media and response selection in a robot call. (a) The public web entry carries the offer and answer. Relayed WebRTC carries speech to Thor and status events back to Safari. (b) Solid arrows trace audio in the response path. Dashed arrows trace the transcript, decision and feedback text. MOSS synthesizes controller feedback as well as chat. In the within-call trace, the commentary run carries an empty turn field, so the bridge infers which feedback item it belongs to. The gate acts on chat audio for the current turn. The session holds the speaker lock during playback, excluding other calls and the local voice process.}\label{fig:architecture}
\end{figure*}

We carry HTTPS signaling through Sakura Frp and relayed WebRTC through Cloudflare TURN \cite{rfc8825,rfc8445,rfc8656,rfc8831}. The media route bypasses the web tunnel. Safari creates a send-only audio transceiver and a data channel for text and status. On the robot, the call backend opens a separate WebSocket to the MOSS speech API. Audio responses go to the room speaker, while the data channel returns transcripts and playback progress. Figure~\ref{fig:architecture} shows the routes and the response-selection path. Each connection exposes its own state for inspection and reconnection.

The gateway and robot still depend on network reachability along their respective routes.

Section~\ref{sec:ready} checks application readiness across the complete path and reports the earlier integration observations.

\subsection{Overlapping connection setup}
We coordinate credential retrieval, microphone permission and relay-candidate gathering before sending an offer. Waiting for each operation to finish before beginning the next would serialize independent work. Candidate gathering also needs an exit condition when one available relay route is sufficient to proceed.

We prefetch TURN credentials when the page opens and reuse that browser request for up to 1 h. On Call, retrieval overlaps microphone authorization. After the first relay candidate, we allow 500 ms for candidates gathered in the same batch, proceeding immediately if gathering completes. At the 20 s fallback, we retain an available relay candidate and reject negotiation when none exists. Backend offer handling has a 55 s deadline, within which speech-service startup and WebRTC-answer generation run concurrently. Failure in either operation cancels its partner and invokes cleanup. TURN credentials last 2 h. Backend reuse requires enough remaining validity for the 1 h call limit plus a 10 min margin.

The setup budget follows the event path from page opening through credential retrieval, microphone authorization, candidate gathering, offer submission, concurrent speech and answer preparation, and readiness. Ready-to-speak is the interface state in which the page first offers the caller the floor. Prefetch may finish before the tap, and retrieval overlaps microphone authorization. The phone logs the tap, microphone readiness, the pre-offer state and ready-to-speak. Backend status samples add connection states. Credential completion, the first relay candidate and the separate backend operations lack individual timing events in these records, so the observed readiness median has no per-stage decomposition. The 500 ms candidate window and 55 s backend deadline constrain their respective stages. They start at different events and cannot be summed into a tap-to-ready budget.

Section~\ref{sec:ready} reports first-dial readiness for this configuration. Its elapsed times include the application path through setup, and its first-attempt counting rule retains failures even when a separately recorded retry connects.

\subsection{Reserving the session and speaker}
A remote call, local voice process or management entry may request the same speaker. We need to settle ownership while the connection is being prepared, before the caller sees readiness. A process-local flag alone leaves another participating process able to request the device independently.

We reserve the session slot during setup and acquire a non-blocking operating-system file lock on the speaker. A second offer encounters HTTP 409 while the first setup or call owns the line. The local voice process obtains the same file lock, extending exclusion across process boundaries. The operating system releases the lock when its holder exits. Hangup names the current session, and the backend compares that identifier before releasing it. Local components enter through a permission-restricted Unix socket and use the same call path. Management operations are restricted to that socket. The setup exception path closes any partially created session and releases its reservation and lock.

Remote access is unavailable while the local voice process holds the speaker lock.

Section~\ref{sec:occupancy} checks rejection during a remote call and local lock ownership, then checks admission after release. Closing a tab and interrupting connectivity exercise cleanup while the caller's view and the backend's state diverge.

\subsection{Audio blocks, half-duplex input and receipts}
A room microphone may capture the robot's own answer and feed it back to recognition. We also need a playback representation that retains the response identity while audio arrives in multiple blocks. Completion must refer to the run that supplied those blocks.

We resample incoming audio to 16 kHz mono PCM16 and accumulate 3,200 B per block, equivalent to 100 ms. While output is queued or playing, we replace uplink blocks with silence, which also drops an attempted interruption during playback. This application-level half-duplex policy operates alongside the browser's echo-cancellation request. The player accepts 48 kHz mono PCM16 with run, turn and chunk metadata and writes queued audio to the output process. It estimates chunk completion from written bytes and the scheduled playback endpoint, then reports run completion after pending chunks finish. Safari displays transcripts and player progress over the data channel, so text may appear before playback finishes.

Section~\ref{sec:ordering} checks source identity, ordering and software completion in replay. Section~\ref{sec:boundaries} examines the association between a run and its feedback owner.

\subsection{A bounded gate for competing answers}
The speech service prepares chat while the controller examines the same final transcript. We need to choose what enters the speaker queue when either component becomes ready first. Waiting indefinitely for the controller would also hold chat indefinitely when the controller stays silent.

We forward the final transcript to the controller bridge and hold chat audio for up to 2 s. A controller claim inside that window discards the buffered chat. An explicit chat decision releases it. Expiry also releases chat, while the controller request continues and later feedback enters playback. Stop requests bypass the bridge's ordinary event queue. We deduplicate final transcripts by turn and discard queued events older than 4 s before delivery. For session-level commentary with an empty turn field, the existing bridge associates a new audio run with the most recent feedback submitted within a 15 s window. Receipts subsequently consult that run mapping. The within-call trace in Section~\ref{sec:boundaries} exercises this association rule.

A controller response arriving after chat release may add another spoken answer, as in the 3 s replay condition.

Section~\ref{sec:ordering} tests the gate deadline with paired on/off schedules. Section~\ref{sec:boundaries} tests feedback ownership at two asynchronous boundaries, locating where the current association rules require a stronger binding.

\subsection{Cleaning up and releasing resources}
We send a heartbeat when the data channel opens and every 10 s thereafter. Every 5 s, the backend checks the 45 s inactivity threshold and 1 h call limit. Data-channel closure, media-track termination and peer-connection failure or closure also invoke cleanup. The close path stops playback, closes speech and media connections and releases the session slot and speaker lock. Closure is serialized and tolerates repeated requests.

\subsection{Implementation scope}
We used one iPhone 17 with Safari over Wi-Fi and cellular access, with the same Thor speaker and public entry. Controlled replay used a simulated controller. Evaluation of tunnel and gateway security lies outside this study.

\section{Reliability Validation}
\subsection{First-dial readiness}\label{sec:ready}
The integration record of 18 November 2026 separately recorded relay candidates at both endpoints, an answer from the speech service and a player report of 268,800 B at 48 kHz, PCM16, mono, equivalent to 2.8 s. It also recorded HTTP 409 for a second offer and an idle backend after hangup.

We tested the connection path of Sections 4.1--4.2 using first-dial readiness. The operator alternated Wi-Fi and cellular calls on the same authorized, warm-start Safari page, holding questions, device and speaker volume fixed. For cellular calls, the operator disabled Wi-Fi in the phone settings. We froze the calibrated rules in \texttt{protocol.json} before collection. After the fourth pair, the operator amended the target to eight calls per network, recording the change from \texttt{frozen.json} to \texttt{frozen-v2.json}.

Of 16 first dials, 15 reached ready-to-speak. Wi-Fi yielded 7/8 with a median of 5.573 s and range 3.731--7.592 s. Cellular yielded 8/8 with a median of 6.7635 s and range 5.911--10.368 s. The frozen first-attempt rule excluded the additional dials for Wi-Fi pairs 1 and 4 from these counts. Table~\ref{tab:readiness} retains the absent readiness event for pair 4 and reports its retry separately.
\begin{table}[tb]
\centering\small
\caption{Seconds from the first tap to ready-to-speak for the 16 planned calls. A dash marks an absent readiness event. Notes also document the question-level timing exclusions.}\label{tab:readiness}
\begin{tabular*}{\columnwidth}{@{\extracolsep{\fill}}lrr@{}}\toprule
Pair & Wi-Fi & Cellular \\\midrule
01 & 4.258 & 7.307 \\
02 & 5.573 & 6.113 \\
03 & 6.204 & 10.368 \\
04 & ---\textsuperscript{a} & 6.220 \\
05 & 3.731 & 5.911 \\
06 & 4.672 & 6.073 \\
07 & 5.787 & 7.909 \\
08 & 7.592 & 7.535 \\\bottomrule
\end{tabular*}
\par\smallskip
\parbox{\columnwidth}{\footnotesize\raggedright\textsuperscript{a}The separate retry reached readiness in 4.654 s. Timing exclusions were Wi-Fi 1, first question (background speech and repeated transcription), Wi-Fi 2, first question (controller unavailability message, with the re-ask excluded), cellular 7, both questions (ambiguous question--reply mapping), and cellular 8, second question (missing first-text event). The second questions in Wi-Fi 1 and 2 contributed timings.}
\end{table}

Among the 15 ready calls, one failed the fixed two-question exchange and 14 were unchecked, comprising six Wi-Fi and eight cellular calls. Zero complete exchanges were confirmed by turn-by-turn review of the room recording. In Wi-Fi pair 2, the controller reported that the language model was unavailable after readiness. The phone logged \texttt{brain.feedback} and displayed the message in the answer area. A later re-ask on the same call yielded an answer, while the original question remained the failed exchange.

For timing, we matched the phone microphone's speech-segment endpoint to the page's first nonempty answer-text event on the same clock, retaining 25 of 30 question slots after the five exclusions in Table~\ref{tab:readiness}. Their median was 3.010 s with range 2.094--4.176 s, including the first cellular question at 12.973 minus 9.803 s, or 3.170 s.

\subsection{Occupancy and recovery under four disturbances}\label{sec:occupancy}
We tested the session reservation, speaker lock and release path of Sections 4.3 and 4.6 in 20 scored scenarios, five per condition. Each ended with a later call on the same line. The busy-line protocol specified an ordinary second offer. We excluded an initial run that displayed a takeover prompt without sending that offer and scored the prescribed repeat.

During an occupied call, the second offer received HTTP 409 in 5/5 checks while the first call completed its current answer. Following hangup, the next call connected in 5/5 checks. With local voice holding the speaker lock, incoming calls were rejected in 5/5 checks and connected after release in 5/5 checks.

Closing the active Safari tab cleared the backend slot and admitted a later call in 5/5 checks. The fourth initial recording contained only idle samples, so we used its retry for that planned slot. For the outage condition, the operator disabled both networks for 60 s with the call tab open. In all five checks, the backend became idle 31.9--36.1 s after its first connected sample while the phone was still offline, and a new call connected after access returned in 5/5 checks.

\subsection{Ordering and suppression at the shared speaker}\label{sec:ordering}
We tested the bounded gate and playback queue of Sections 4.4--4.5 with paired gate-on and gate-off fixtures. Relative to injection of the final transcript, chat audio arrived at 0.20, 0.45, 0.70, 1.10 or 2.25 s. Controller decisions were scheduled at 1 s for fast tasks and at 3 s for slow tasks, against the 2 s gate. The replay exercised the session, bridge, gate and player through their event interfaces, routing playback to a local sink.

We identified distinct outputs by the player's run identifier and checked their source against \texttt{expected\_by\_run} and \texttt{fixture\_audio} in each trace. Table~\ref{tab:replay} reports all 40 schedules. With the gate enabled, chat was suppressed for every fast-task timing. At 3 s both settings yielded chat followed by controller feedback. The chat-decision and no-decision conditions started the same chat source, while decision events distinguished explicit release from timeout.
\begin{table*}[t]
\centering\small
\caption{Software outputs by source. A count of 5/5 represents one trace at each of five audio timings. The middle column reports traces with two outputs, gate off / on.}\label{tab:replay}
\begin{tabularx}{\linewidth}{@{}p{.23\linewidth}p{.18\linewidth}X@{}}\toprule
Controller condition & Two outputs\newline off / on & Surviving output identity \\\midrule
Chat decision & 0/5 / 0/5 & Both settings: chat only. \\
No decision & 0/5 / 0/5 & Both settings: chat only. \\
Fast task, 1 s & 5/5 / 0/5 & Off: chat and controller feedback. On: controller feedback only. \\
Slow task, 3 s & 5/5 / 5/5 & Both settings: chat followed by controller feedback. \\\bottomrule
\end{tabularx}
\end{table*}
\begin{figure*}[t]
\centering
\includegraphics[width=.98\linewidth]{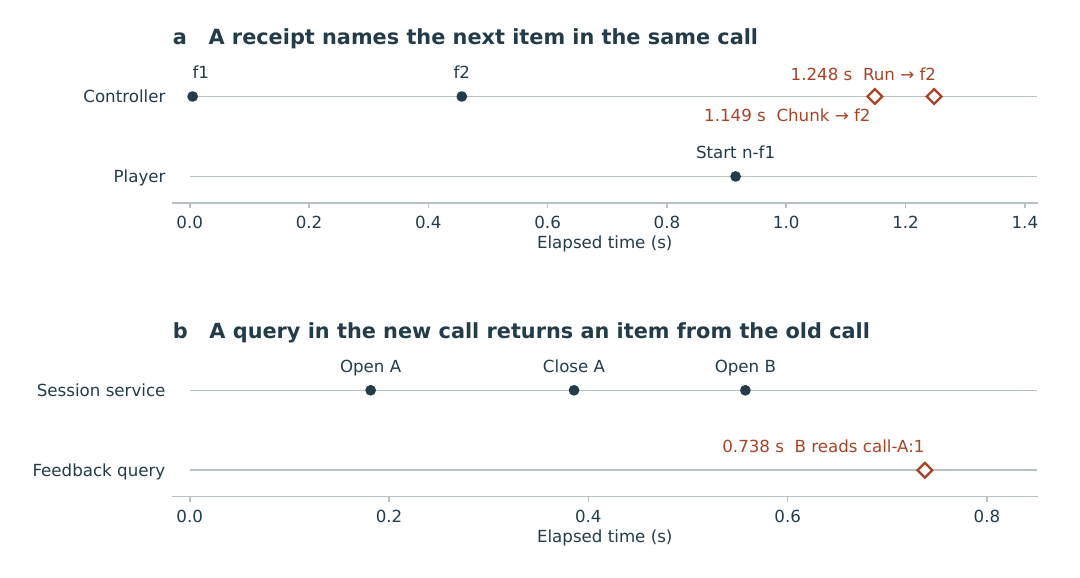}
\caption{Two boundary traces with independent time origins. Filled points mark logged events. Open diamonds mark reports whose assigned owner differs from the originating item. In (a), time starts at playback-fixture construction. The player starts run \texttt{n-f1}, but the controller receives both its first chunk receipt and its run receipt under \texttt{f2}. In (b), time starts just before the test service is started. Points mark the return of each session or query request. The trace ends with the query result.}\label{fig:boundary}
\end{figure*}
We checked player start and finish intervals for distinct runs and the feedback identifier on completion receipts. All 40 scheduled traces had no overlapping output and matching feedback identifiers. With the fast-task gate disabled, the controller reached playback first at the latest chat arrival setting, showing that the queue preserved arrival-dependent order.

\subsection{Two remaining errors at asynchronous boundaries}\label{sec:boundaries}
We tested the feedback-association rule of Section 4.5 at boundaries where more than one item could supply an owner. The two fixtures were collected separately from the 40 schedules. Figure~\ref{fig:boundary} retains their independent clocks.

\noindent Within a call, Figure~\ref{fig:boundary}a uses the monotonic clock at playback-fixture construction as time zero, before either feedback submission. The fixture supplied \texttt{f1} at 0.005 s and \texttt{f2} at 0.456 s. Its reference mapping assigns the audio run \texttt{n-f1} to \texttt{f1}. The arriving audio carried that run identifier and an empty \texttt{turn\_id}.

For an unseen run with an empty turn, the bridge assigns the latest feedback submitted within 15 s and uses that mapping for subsequent receipts. By the arrival of \texttt{n-f1}, the remembered submission was \texttt{f2}.

The player began \texttt{n-f1} at 0.915 s. At 1.149 s the controller-side log recorded the first chunk completion for that run, and at 1.248 s it recorded completion of the whole run. Both reports carried \texttt{feedback\_id=f2}. Both retained the correct call identifier, \texttt{call-9e49673f268254bd07e396f5}. The lost distinction was between feedback items within that call.

\begin{table*}[!t]
\centering\small
\caption{Design choices, addressed failures and supporting observations.}\label{tab:mapping}
\begin{tabularx}{\linewidth}{@{}p{.28\linewidth}p{.30\linewidth}X@{}}\toprule
Design choice & Addressed failure & Evidence \\\midrule
Separate signaling, media and speech routes & A page-load event standing in for a working call & Integration record of 18 November (Section 5.1). \\
Credential prefetch, 500 ms candidate window and 55 s concurrent setup & Serial setup waits and first-dial failure & 15/16 ready, with medians 5.573 s and 6.7635 s (Section 5.1). \\
Session slot and non-blocking speaker lock & Multiple calls or a local process writing to one speaker & 20/20 later calls connected, HTTP 409 in 5/5 busy-line checks (Section 5.2). \\
Half-duplex uplink during playback & The robot's output re-entering recognition & Implementation policy in Section 4.4. Acoustic effect unmeasured. \\
2 s gate, timeout release and stop bypass & Competing answers entering playback & 40 traces without overlapping output, chat suppressed 5/5 at 1 s with the gate enabled (Section 5.3). Stop bypass unmeasured separately. \\
Run/turn/chunk binding on receipts (proposed repair) & A receipt or late feedback acquiring the wrong owner & Two association errors motivate the binding and generation checks (Section 5.4). \\\bottomrule
\end{tabularx}
\par\smallskip
\parbox{\linewidth}{\footnotesize\raggedright The observations support the listed behavior under the stated conditions. The final row specifies a repair motivated by the traces.}
\end{table*}

\noindent Across calls, Figure~\ref{fig:boundary}b sets a fresh monotonic origin just before feedback-service startup and records each timestamp after the corresponding request returns. The service acknowledged opening \texttt{call-A} at 0.182 s, closing A at 0.386 s and opening \texttt{call-B} at 0.558 s. Only then did the fixture submit the delayed completion of \texttt{task-A} with the original request identifier \texttt{call-A:1}.

At 0.738 s the feedback query for B returned that item. Its \texttt{session\_id} named \texttt{call-B}, while both \texttt{request\_id} and \texttt{event\_id} retained \texttt{call-A:1}. The request and event fields still identify A's turn. This timestamp marks receipt of the query result.

The fixtures establish a mismatched receipt owner within one call and delivery into a later call's feedback query.

\subsection{Design choices and their evidence}\label{sec:mapping}
We matched each mechanism to the failure boundary it addresses and the observation used to inspect it. Table~\ref{tab:mapping} includes the proposed receipt-binding repair as a design implication of the boundary traces.

\section{Discussion}
\subsection{Why the boundaries matter}
We place ownership decisions at resource boundaries. The setup reservation covers the interval before readiness, and the file lock extends exclusion across participating processes. After gate expiry, the continuing controller request explains the second sequential answer in the slow condition. Resource admission and turn selection operate on different lifetimes.

Wi-Fi pair 2 makes the separation between layers concrete. The line reached readiness, then the controller returned a language-model-unavailable message that appeared in the answer area. Recording \texttt{brain.feedback} alongside readiness preserves both the working channel and the failed exchange. A call interface could use those sources to show progress at the corresponding layer.

\Needspace{5\baselineskip}
Release also has a source and a clock. The outage intervals were 8.9--13.1 s shorter than the configured 45 s inactivity threshold. Status samples bracketing each session also bound its entire occupancy below 45 s. A 45 s inactivity timeout therefore cannot have completed within that session, excluding the heartbeat path. Section 4.6 lists three further triggers, media-track termination, peer-connection failure or closure, and data-channel closure. The logs do not identify which of these fired. Redundant release paths make the call resilient to any single signal failing, at the cost of not knowing afterwards which one acted. Backend release while the phone remains offline gives the returning interface a state to reconcile. A release event carrying the call identifier and cause would let it do so.

\subsection{The costs of waiting and half-duplex input}
We trade waiting time against the opportunity for a late controller answer to follow chat. Spoken-dialogue work relates response timing to the transfer of the floor \cite{skantze2014,skantze2021}. Here, the floor is available but its software owner is undecided. Extending the gate would extend the delay whenever the controller stays silent. Shortening it would bring chat release closer to controller deliberation. The tested deadline defines which of the scheduled decisions suppresses chat.

Half-duplex input introduces another explicit cost. Replacing uplink blocks with silence prevents caller speech during playback from reaching the service, including an attempted spoken interruption in that interval. The stop-event bypass concerns delivery after a stop request has been recognized. It addresses queue waiting at that later stage. Local speaker ownership similarly makes the public entry unavailable for the duration of local use. These are operational consequences of the mechanisms that an interface needs to expose.

\subsection{Transferable mechanisms and deployment dependencies}
We can carry the separation of connection states, process-level exclusion and bounded turn arbitration into another voice path by identifying its corresponding resources and events. Telepresence and cloud-robotics work situate remote access across physical and computational boundaries \cite{kristoffersson,kehoe2015,toris}. Our implementation adds explicit admission and output decisions at the shared device, using ordinary WebRTC and operating-system facilities.

The concrete path depends on Safari's media behavior, TURN access, the tunnel and the MOSS event schema. WebRTC transports the application identifiers attached to a message \cite{rfc8825,rfc8831}. A receipt's owner therefore remains an application responsibility. The within-call trace motivates a stable call, turn and chunk binding carried from feedback submission into the audio run and its completion reports. The cross-call trace motivates checking the originating call and connection generation when feedback is stored and again when delivered.

\section{Limitations}
The field calls used one device and browser, and the target of eight calls per network was set after the fourth pair. Fourteen ready first dials lack a checked exchange outcome. Timing ends at answer text on the phone, leaving audible-response time unmeasured. Replay used a simulated controller and a local sink at the listed schedules. The boundary fixtures locate association errors without estimating their deployed frequency. We evaluated these implementation behaviors without a baseline system or a user study.

\section{Conclusion}
We put a robot on a browser voice call through separate signaling, relayed-media and speech-service paths. Prefetch and concurrent setup supported 15/16 ready first dials in the recorded configuration. Session reservation and a shared speaker lock governed admission across remote and local use, with later calls admitted in all 20 scored scenarios. A 2 s gate suppressed chat for the 1 s controller decisions and released it before the 3 s decisions, leaving sequential answers. The two boundary traces connect the remaining feedback errors to identifier scope and motivate preserving the originating call and turn through asynchronous delivery.

\bibliography{references}

\begin{thebibliography}{26}
\providecommand{\natexlab}[1]{#1}
\providecommand{\url}[1]{{#1}}
\providecommand{\urlprefix}{URL }
\providecommand{\doi}[1]{\url{https://doi.org/#1}}
\providecommand{\eprint}[2][]{\url{#2}}
 \bibcommenthead

\bibitem[{Alvestrand(2021)}]{rfc8825}
Alvestrand H (2021) {Overview: Real-Time Protocols for Browser-Based
  Applications}. Tech. Rep. RFC 8825, Internet Engineering Task Force,
  \doi{10.17487/RFC8825},
  \urlprefix\url{https://www.rfc-editor.org/rfc/rfc8825}

\bibitem[{Bangor et~al.(2008)Bangor, Kortum, and Miller}]{bangor2008}
Bangor A, Kortum PT, Miller JT (2008) {An Empirical Evaluation of the System
  Usability Scale}. International Journal of Human--Computer Interaction
  24(6):574--594. \doi{10.1080/10447310802205776},
  \urlprefix\url{https://doi.org/10.1080/10447310802205776}

\bibitem[{Bartneck et~al.(2009)Bartneck, Kuli{\'c}, Croft, and
  Zoghbi}]{bartneck2009}
Bartneck C, Kuli{\'c} D, Croft E, et~al (2009) {Measurement Instruments for the
  Anthropomorphism, Animacy, Likeability, Perceived Intelligence, and Perceived
  Safety of Robots}. International Journal of Social Robotics 1(1):71--81.
  \doi{10.1007/s12369-008-0001-3},
  \urlprefix\url{https://doi.org/10.1007/s12369-008-0001-3}

\bibitem[{Bohus and Horvitz(2009)}]{bohus2009}
Bohus D, Horvitz E (2009) {Models for multiparty engagement in open-world
  dialog}. In: Proceedings of the SIGDIAL 2009 Conference on The 10th Annual
  Meeting of the Special Interest Group on Discourse and Dialogue - SIGDIAL
  '09, pp 225--234, \doi{10.3115/1708376.1708409},
  \urlprefix\url{https://doi.org/10.3115/1708376.1708409}

\bibitem[{Bohus and Horvitz(2014)}]{bohus2014}
Bohus D, Horvitz E (2014) {Managing Human-Robot Engagement with Forecasts
  and... um ... Hesitations}. In: Proceedings of the 16th International
  Conference on Multimodal Interaction, pp 2--9, \doi{10.1145/2663204.2663241},
  \urlprefix\url{https://doi.org/10.1145/2663204.2663241}

\bibitem[{Cherakara et~al.(2023)Cherakara, Varghese, Shabana, Nelson,
  Karukayil, Kulothungan, Afil~Farhan, Nesset, Moujahid, Dinkar, Rieser, and
  Lemon}]{furchat2023}
Cherakara N, Varghese F, Shabana S, et~al (2023) {FurChat: An Embodied
  Conversational Agent using LLMs, Combining Open and Closed-Domain Dialogue
  with Facial Expressions}. In: Proceedings of the 24th Meeting of the Special
  Interest Group on Discourse and Dialogue, pp 588--592,
  \doi{10.18653/v1/2023.sigdial-1.55},
  \urlprefix\url{https://doi.org/10.18653/v1/2023.sigdial-1.55}

\bibitem[{Clark et~al.(2019)Clark, Pantidi, Cooney, Doyle, Garaialde, Edwards,
  Spillane, Gilmartin, Murad, Munteanu, Wade, and Cowan}]{clark2019}
Clark L, Pantidi N, Cooney O, et~al (2019) {What Makes a Good Conversation?
  Challenges in Designing Truly Conversational Agents}. In: Proceedings of the
  2019 CHI Conference on Human Factors in Computing Systems, pp 1--12,
  \doi{10.1145/3290605.3300705},
  \urlprefix\url{https://doi.org/10.1145/3290605.3300705}

\bibitem[{Desai et~al.(2011)Desai, Tsui, Yanco, and Uhlik}]{desai2011}
Desai M, Tsui KM, Yanco HA, et~al (2011) {Essential features of telepresence
  robots}. In: 2011 IEEE Conference on Technologies for Practical Robot
  Applications, pp 15--20, \doi{10.1109/tepra.2011.5753474},
  \urlprefix\url{https://doi.org/10.1109/tepra.2011.5753474}

\bibitem[{Fong et~al.(2003)Fong, Nourbakhsh, and Dautenhahn}]{fong2003}
Fong T, Nourbakhsh I, Dautenhahn K (2003) {A survey of socially interactive
  robots}. Robotics and Autonomous Systems 42(3-4):143--166.
  \doi{10.1016/s0921-8890(02)00372-x},
  \urlprefix\url{https://doi.org/10.1016/s0921-8890(02)00372-x}

\bibitem[{Goodrich and Schultz(2008)}]{goodrich}
Goodrich MA, Schultz AC (2008) {Human--Robot Interaction: A Survey}.
  Foundations and Trends in Human--Computer Interaction 1(3):203--275.
  \doi{10.1561/1100000005}, \urlprefix\url{https://doi.org/10.1561/1100000005}

\bibitem[{Jesup et~al.(2021)Jesup, Loreto, and T{\"u}xen}]{rfc8831}
Jesup R, Loreto S, T{\"u}xen M (2021) {WebRTC Data Channels}. Tech. Rep. RFC
  8831, Internet Engineering Task Force, \doi{10.17487/RFC8831},
  \urlprefix\url{https://www.rfc-editor.org/rfc/rfc8831}

\bibitem[{Kehoe et~al.(2015)Kehoe, Patil, Abbeel, and Goldberg}]{kehoe2015}
Kehoe B, Patil S, Abbeel P, et~al (2015) {A Survey of Research on Cloud
  Robotics and Automation}. IEEE Transactions on Automation Science and
  Engineering 12(2):398--409. \doi{10.1109/tase.2014.2376492},
  \urlprefix\url{https://doi.org/10.1109/tase.2014.2376492}

\bibitem[{Ker{\"a}nen et~al.(2018)Ker{\"a}nen, Holmberg, and
  Rosenberg}]{rfc8445}
Ker{\"a}nen A, Holmberg C, Rosenberg J (2018) {Interactive Connectivity
  Establishment (ICE): A Protocol for Network Address Translator (NAT)
  Traversal}. Tech. Rep. RFC 8445, Internet Engineering Task Force,
  \doi{10.17487/RFC8445},
  \urlprefix\url{https://www.rfc-editor.org/rfc/rfc8445}

\bibitem[{Kidd and Breazeal(2004)}]{kidd2004}
Kidd C, Breazeal C (2004) {Effect of a robot on user perceptions}. In: 2004
  IEEE/RSJ International Conference on Intelligent Robots and Systems (IROS)
  (IEEE Cat. No.04CH37566), pp 3559--3564, \doi{10.1109/iros.2004.1389967},
  \urlprefix\url{https://doi.org/10.1109/iros.2004.1389967}

\bibitem[{Kristoffersson et~al.(2013)Kristoffersson, Coradeschi, and
  Loutfi}]{kristoffersson}
Kristoffersson A, Coradeschi S, Loutfi A (2013) {A Review of Mobile Robotic
  Telepresence}. Advances in Human-Computer Interaction 2013:1--17.
  \doi{10.1155/2013/902316},
  \urlprefix\url{https://doi.org/10.1155/2013/902316}

\bibitem[{Li(2015)}]{li2015}
Li J (2015) {The benefit of being physically present: A survey of experimental
  works comparing copresent robots, telepresent robots and virtual agents}.
  International Journal of Human-Computer Studies 77:23--37.
  \doi{10.1016/j.ijhcs.2015.01.001},
  \urlprefix\url{https://doi.org/10.1016/j.ijhcs.2015.01.001}

\bibitem[{Luger and Sellen(2016)}]{luger2016}
Luger E, Sellen A (2016) {"Like Having a Really Bad PA": The Gulf between User
  Expectation and Experience of Conversational Agents}. In: Proceedings of the
  2016 CHI Conference on Human Factors in Computing Systems, pp 5286--5297,
  \doi{10.1145/2858036.2858288},
  \urlprefix\url{https://doi.org/10.1145/2858036.2858288}

\bibitem[{Porcheron et~al.(2018)Porcheron, Fischer, Reeves, and
  Sharples}]{porcheron2018}
Porcheron M, Fischer JE, Reeves S, et~al (2018) {Voice Interfaces in Everyday
  Life}. In: Proceedings of the 2018 CHI Conference on Human Factors in
  Computing Systems, pp 1--12, \doi{10.1145/3173574.3174214},
  \urlprefix\url{https://doi.org/10.1145/3173574.3174214}

\bibitem[{Reddy et~al.(2020)Reddy, Johnston, Matthews, and Rosenberg}]{rfc8656}
Reddy T, Johnston A, Matthews P, et~al (2020) {Traversal Using Relays around
  NAT (TURN): Relay Extensions to Session Traversal Utilities for NAT (STUN)}.
  Tech. Rep. RFC 8656, Internet Engineering Task Force, \doi{10.17487/RFC8656},
  \urlprefix\url{https://www.rfc-editor.org/rfc/rfc8656}

\bibitem[{Sacks et~al.(1974)Sacks, Schegloff, and Jefferson}]{sacks1974}
Sacks H, Schegloff EA, Jefferson G (1974) {A simplest systematics for the
  organization of turn-taking for conversation}. Language 50(4):696--735.
  \doi{10.2307/412243}, \urlprefix\url{https://doi.org/10.2307/412243}

\bibitem[{Skantze(2021)}]{skantze2021}
Skantze G (2021) {Turn-taking in Conversational Systems and Human-Robot
  Interaction: A Review}. Computer Speech \& Language 67:101178.
  \doi{10.1016/j.csl.2020.101178},
  \urlprefix\url{https://doi.org/10.1016/j.csl.2020.101178}

\bibitem[{Skantze et~al.(2014)Skantze, Hjalmarsson, and Oertel}]{skantze2014}
Skantze G, Hjalmarsson A, Oertel C (2014) {Turn-taking, feedback and joint
  attention in situated human--robot interaction}. Speech Communication
  65:50--66. \doi{10.1016/j.specom.2014.05.005},
  \urlprefix\url{https://doi.org/10.1016/j.specom.2014.05.005}

\bibitem[{Steinfeld et~al.(2006)Steinfeld, Fong, Kaber, Lewis, Scholtz,
  Schultz, and Goodrich}]{steinfeld}
Steinfeld A, Fong T, Kaber D, et~al (2006) {Common metrics for human-robot
  interaction}. In: Proceedings of the 1st ACM SIGCHI/SIGART conference on
  Human-robot interaction, pp 33--40, \doi{10.1145/1121241.1121249},
  \urlprefix\url{https://doi.org/10.1145/1121241.1121249}

\bibitem[{Tellex et~al.(2020)Tellex, Gopalan, Kress-Gazit, and
  Matuszek}]{tellex2020}
Tellex S, Gopalan N, Kress-Gazit H, et~al (2020) {Robots That Use Language}.
  Annual Review of Control, Robotics, and Autonomous Systems 3:25--55.
  \doi{10.1146/annurev-control-101119-071628},
  \urlprefix\url{https://doi.org/10.1146/annurev-control-101119-071628}

\bibitem[{Toris et~al.(2015)Toris, Kammerl, Lu, Lee, Jenkins, Osentoski, Wills,
  and Chernova}]{toris}
Toris R, Kammerl J, Lu DV, et~al (2015) {Robot Web Tools: Efficient messaging
  for cloud robotics}. In: 2015 IEEE/RSJ International Conference on
  Intelligent Robots and Systems (IROS), pp 4530--4537,
  \doi{10.1109/iros.2015.7354021},
  \urlprefix\url{https://doi.org/10.1109/iros.2015.7354021}

\bibitem[{Tsui et~al.(2011)Tsui, Desai, Yanco, and Uhlik}]{tsui2011}
Tsui KM, Desai M, Yanco HA, et~al (2011) {Exploring use cases for telepresence
  robots}. In: Proceedings of the 6th international conference on Human-robot
  interaction, pp 11--18, \doi{10.1145/1957656.1957664},
  \urlprefix\url{https://doi.org/10.1145/1957656.1957664}

\end{thebibliography}
\end{document}